\documentclass{article}

\usepackage[main, preprint]{neurips_2026}

\title{Eliciting associations between clinical variables from LLMs via comparison questions across populations}
\author{%
  \parbox{0.97\textwidth}{\centering
    Fabian Kabus$^{1}$, Kian Kordtomeikel$^{2}$, Thomas Brox$^{2}$, Heinz Wiendl$^{4}$,\\
    Daiana Stolz$^{3}$, Harald Binder$^{1}$\\[0.55em]
    {\mdseries
    $^{1}$Institute of Medical Biometry and Statistics (IMBI), Medical Center, University of Freiburg\\
    $^{2}$Department of Computer Science, Faculty of Engineering, University of Freiburg\\
    $^{3}$Department of Pneumology, Medical Center, University of Freiburg\\
    $^{4}$Department of Neurology and Neurophysiology, Medical Center, University of Freiburg\\[0.75em]
    \texttt{fabian.kabus@uniklinik-freiburg.de}\\
    }
  }
}

\usepackage{amsmath,amssymb}
\usepackage{graphicx}
\usepackage[table]{xcolor}
\usepackage{tabularray}
\UseTblrLibrary{booktabs}
\usepackage{booktabs}
\usepackage{tabularx}
\usepackage{subcaption}
\usepackage{placeins}
\usepackage{url}

\DeclareMathOperator{\Var}{Var}

\newcommand{\vname}[1]{#1}

\begin{document}

\maketitle

\begin{abstract}
The training data of large language models (LLMs) comprises a wide range of biomedical literature, reflecting data from many different patient populations. We investigate how it might be possible to recover information on correlation and causal links between patient characteristics, as a key building block for medical decision making. To avoid the pitfalls of 
direct elicitation, we propose an approach based on structured comparison questions, specifically patient comparison triplet questions. This is combined with a statistical model for the LLM representation that provides estimates of correlations without access to activations or model internals. Intuitively, we consider how similarity decisions of LLMs based on a first variable are affected by providing information on a second variable for one of the patients being assessed. 
We then induce prompt-level environment shifts to obtain correlation estimates for different subpopulations, which enables an invariant causal prediction (ICP) approach to obtain conservative candidate parent links.
We demonstrate the method in two clinical domains, chronic obstructive pulmonary disease (COPD) and multiple sclerosis (MS).
Across prompted environments, the elicited correlations are smooth, stable, and clinically interpretable, yet vary in a statistically significant way that supports downstream invariance testing, such that ICP provides a small set of candidate invariant parent links.
These results show that indirect elicitation via triplet comparisons can recover meaningful association structure from LLMs and offer a cautious route from implicit correlations to causal statements that are congruent with LLM answering patterns.
\end{abstract}


\section{Introduction}

As large language models (LLMs) are becoming a viable option for clinical decision tasks \citep{brodeur2026llm_reasoning}, a major question is how to best query LLMs, and how the representation of clinical settings by LLMs could be formalized. We focus on the task of assessing the association between different patient characteristics, via correlation or causal links, as a key ingredient of clinical decisions. As LLMs have been trained on a multitude of biomedical papers describing a broad range of different patient populations, it might be feasible to not only obtain a general statement, e.g. on the correlation of two clinical markers, but values specific to a decision task and subpopulation at hand, by getting the LLM to give more weight to relevant evidence.
An obvious approach for retrieving such information is to ask directly: "What is the correlation between X and Y in population Z?"
Direct queries, however, may fail in multiple ways.
First, the answer may reflect recall of a specific statement rather than aggregated knowledge across studies.
Second, even when the model responds with a plausible number, we cannot verify whether it respects that relationship when subsequently used for decision-making.
Beyond these fundamental issues, models exhibit sycophancy by adjusting answers to match perceived user beliefs \citep{sharma2025understanding}, responses are sensitive to the order of options \citep{pezeshkpour2023large}, outputs may be hallucinated without grounding \citep{ji2023survey}, and chain-of-thought explanations can reflect post-hoc rationalization rather than actual computation \citep{lindsey2025biology}.
This implies that when extracting LLM knowledge, we should treat LLMs as experimental subjects to be measured, not as self-reflecting agents to be asked.

The cognitive sciences have faced similar challenges when investigating the representations of human subjects, and have developed approaches for obtaining quantitative representations from structured comparison questions \citep{suppes1989foundations}. Inspired by this research, we specifically propose to query LLMs with triplet questions---``Is entity 3 more similar to entity 1 or to entity 2?''---to obtain implicit correlations between patient characteristics, and to formalize the problem representation of LLMs. 
This also is related to more recent results, where triplet questions have been used to recover similarity structure \citep{vankadara2023insights} and applied to derive clinical concept embeddings from LLMs \citep{kabus2026assessing}.

To obtain information about the relationship between two patient characteristics, i.e. variables, we construct a triplet question involving three artificial patients, denoted 1, 2, and 3. We specify the values of the first variable for all three patients, but a value for the second variable is specified only for patient 3. This allows us to observe whether the value of the second variable tips the balance of the LLM similarity assessment of patient 3. By systematically varying the values, we can then assess how extreme the values of the second variable have to be to influence the decision, which carries information on correlation. We formalize this intuition into a statistical surrogate model for the LLM representation and decision to obtain estimates of implied correlation coefficients, including confidence intervals.
We demonstrate the method in two clinical settings, with lung function markers in chronic obstructive pulmonary disease (COPD) and multiple sclerosis (MS) biomarkers, highlighting how different correlation estimates are obtained for different patient populations, contrasted with results from directly requesting correlation values from LLMs.

Having access to estimated correlations from different subpopulations is also useful for causal assessment. 
Specifically, Invariant Causal Prediction \citep[ICP;][]{peters2016causal} allows us to test for causal links when given observational data from multiple environments, such as patient populations, that share the same causal structure.
The invariance principle is as follows: if $X$ causally affects $Y$, then the conditional distribution $P(Y \mid X)$, e.g. as assessed by regression models, remains stable across environments, even as the marginal distribution of $X$ shifts.
Spurious correlations, by contrast, break under distribution shift. The original ICP approach requires several individual-level datasets from different patient populations, but these might be difficult to obtain in practice. We demonstrate how the required regression parameter estimates can alternatively be obtained from our proposed correlation estimations, including uncertainty to enable statistical testing.
To implement this, we synthesize environments through prompt variations, preceding the triplet question, by specifying the mean of the variable in the subpopulation under consideration.
This allows us to identify causal relationships that congruently describe LLM decision behavior.

Our contributions are as follows:
\begin{itemize}
    \item \textbf{Triplet-based extraction of clinical associations.}
    We introduce a framework that recovers pairwise correlations from LLM similarity judgments without access to model internals.
    \item \textbf{Prompted environments for causal assessment.}
    We estimate environment-specific correlations in prompted subpopulations and use ICP to obtain candidate parent links.
    \item \textbf{Clinical demonstration in COPD and MS.}
    We show that the extracted associations are clinically interpretable, more stable than direct correlation queries, and support a small set of plausible invariant parent links.
\end{itemize}

Empirically, triplet decision surfaces are smooth and well fit by the surrogate model, and triplet-based correlation estimates are substantially less noisy than direct numeric queries.
The implied correlations are clinically plausible and generally align known characteristics.
Further, across prompted environments, ICP identifies a small conservative set of invariant parent links.

Section~\ref{sec:methods} introduces the triplet representation and decision model, derives the symmetric correlation estimator, and describes the uncertainty-aware ICP procedure with prompted environment shifts.
Section~\ref{sec:experiments} describes the clinical applications and experimental protocol, Section~\ref{sec:results} reports the empirical results, Section~\ref{sec:related_work} discusses related work, and Section~\ref{sec:conclusion} provides concluding remarks.

\section{Methods}
\label{sec:methods}

Our method treats the LLM as an experimental subject from which information on the association between different variables, e.g., different patient characteristics, is to be obtained. The basic building block is the elicitation of correlation for pairs of variables. 
For each variable pair, we ask patient comparison triplet questions, fit a surrogate model to the resulting binary choices, and recover the model's implied correlation from the fitted slopes.
Repeating this procedure for different prompted environments (i.e. patient subpopulations) yields correlation matrices that we pass to invariant causal prediction.

\subsection{Triplet questions}
\label{sec:triplet_queries}

Let $\mathcal{X} = \{X_1, \dots, X_p\}$ denote the variable set of interest.
To extract pairwise correlation structure, we query the LLM about three patients, Patient 1, 2, and 3, and pairs of variables $X_j$ and $X_k$, $j \neq k$. 
All experiments are replicated across distinct environments $e \in \{1, \dots, m\}$, where each environment has its own implied population means $\mu^{(e)} = \left(\mathbb{E}_e[X_1],\ldots,\mathbb{E}_e[X_p]\right)$, and is implemented via a corresponding prompt text.
Variation in these means serves as a synthetic distribution shift for the downstream ICP analysis.

For a fixed environment $e$, $X_j$ is reported for all three patients, i.e. 
the question specifies values $X_j^{(1)}, X_j^{(2)}, X_j^{(3)}$. The value of the other variable $X_k^{(3)}$ is specified only for Patient 3.
The LLM is asked whether Patient 3 is more similar to Patient 1 or to Patient 2, taking $X_k^{(3)}$ into account.
The binary response $Y_{jk}^{(e)} \in \{1, 2\}$ is our observation for subsequent modeling. 
We collect responses per environment by varying $(X_j^{(3)}, X_k^{(3)})$ over a regular grid while holding $(X_j^{(1)}, X_j^{(2)})$ fixed. If the representation of the LLM entails a strong relation between $X_j$ and $X_k$, presentation of a specific value $X_k^{(3)}$ in relation to the values $X_j^{(1)}, X_j^{(2)}, X_j^{(3)}$ intuitively should more strongly influence the answer. We also consider a version of the questions, where the roles of $X_j$ and $X_k$ are flipped. 
The symmetric estimator of the correlation in Section~\ref{sec:correlation_estimation} integrates LLM answers from both versions.

\subsection{Decision surrogate model based on a rational LLM representation}
\label{sec:decision_surrogate}

For inferring a correlation from triplet question answers, we assume an LLM representation that reflects rational behavior. Specifically, the value $X_j^{(3)}$ is assumed to be a noisy measurement of the true state $X_j^{*,(3)}$ of Patient 3, i.e. $X_j^{(3)} = X_j^{*,(3)} + \epsilon_j$, with $\mathbb{E}(\epsilon_j) = 0$ and $\Var(\epsilon_j) = \sigma_j^2$. The additionally specified value $X_k^{(3)}$ then provides a second route towards $X_j^{*,(3)}$. We assume a linear relation $X_j^* = a_0 + a_1 X_k + U_{j|k}$,
where $U_{j|k} \perp X_k$ with variance $\tau_{j|k}^2$.
Letting $s_j^2 = \Var(X_j)$ and $s_k^2 = \Var(X_k)$, the slope is directly related to the population correlation $\rho_{jk}$ via 
\begin{equation}
\label{eq:acor}
a_1 = \rho_{jk} s_j / s_k.
\end{equation}

Assuming the LLM representation entails a measurement error perspective, the LLM prediction would incorporate the projection
\begin{equation}
\label{eq:projection}
X_{j|k}^{(3)} = a_0 + a_1 X_k^{(3)},
\end{equation}
i.e. the value of $X_j^{*,(3)}$ predicted from $X_k^{(3)}$ for minimizing measurement error. Its residual variance is $\tau_{j|k}^2 = s_j^2(1 - \rho_{jk}^2)$, and the total error variance with respect to 
$X_j^{*,(3)}$ is $v_{j|k}^2 = \sigma_j^2 + \tau_{j|k}^2$.

The statistically optimal combination of the direct measurement $X_j^{(3)}$ and the indirect measurement $X_{j|k}^{(3)}$ for assessing $X_j^{*,(3)}$ is the inverse-variance estimate
\begin{equation}
\label{eq:invvar}
    \hat{X}_j^{*,(3)} = w_1 X_j^{(3)} + w_2 X_{j|k}^{(3)}, \qquad w_1 = \frac{v_{j|k}^2}{\sigma_j^2 + v_{j|k}^2}, \quad w_2 = \frac{\sigma_j^2}{\sigma_j^2 + v_{j|k}^2}.
\end{equation}
For providing an answer based on this estimate, the LLM ideally should compare $\hat{X}_j^{*,(3)}$ to the reference midpoint $X_{j,\mathrm{ref}} = (X_j^{(1)}+X_j^{(2)})/2$, similar to a regression model

\begin{eqnarray} 
    P\!\left(Y_{jk}^{(e)} = 2 \mid X_j^{(3)}, X_k^{(3)}\right)
    &=& h\left(\beta_s\cdot(\hat{X}_j^{*,(3)} - X_{j,\mathrm{ref}})\right)\\
\label{eq:logistic_surrogate}
    &=& 
    h\left(\beta_{0,jk}^{(e)} + \beta_{1,jk}^{(e)} X_j^{(3)} + \beta_{2,jk}^{(e)} X_k^{(3)}\right)
\end{eqnarray}
with $h(\eta)=1/(1+\exp(-\eta))$, and scaling parameter $\beta_s$. 
Expanding (\ref{eq:logistic_surrogate}) using (\ref{eq:projection}) and (\ref{eq:invvar}) shows that $\beta_{1,jk}^{(e)} = \beta_s w_1$ and $\beta_{2,jk}^{(e)} = \beta_s w_2 a_1 = \beta_s w_2 \rho_{jk}\frac{s_j}{s_k}$, hence
\begin{equation}
\label{eq:directional_ratio}
    \frac{\beta_{2,jk}^{(e)}}{\beta_{1,jk}^{(e)}}
    =
    \frac{w_2}{w_1}\frac{s_j}{s_k}\rho_{jk},
\end{equation}
tying $\rho_{jk}$ to the fitted logistic coefficients $\beta_{1,jk}^{(e)}$ and $\beta_{2,jk}^{(e)}$ through their slope ratio.

\subsection{Correlation estimation}
\label{sec:correlation_estimation}

\paragraph{Symmetric estimator.}
The directional relation in (\ref{eq:directional_ratio}) depends not only on the correlation $\rho_{jk}$, but also on nuisance terms tied to variable scaling and relative cue weighting. Using the alternative question version, which flips the role of $X_j$ and $X_k$, we obtain an estimate of $\rho_{kj}$. 
We can exploit two symmetry properties: first, the true correlation is symmetric, $\rho_{jk} = \rho_{kj}$, so both question versions target the same quantity; second, reversing the query exchanges the roles of the two variables and of the corresponding evidence channels.
Under this paired-query symmetry, the $s_j/s_k$ scaling cancels exactly and the remaining directional asymmetry is reduced in the product of the two slope ratios 
\begin{equation} \label{eq:slope_product}
    \frac{\hat{\beta}_{2,jk}^{(e)}}{\hat{\beta}_{1,jk}^{(e)}}
    \cdot
    \frac{\hat{\beta}_{2,kj}^{(e)}}{\hat{\beta}_{1,kj}^{(e)}}
    \approx
    \left(\rho_{jk}^{(e)}\right)^2.
\end{equation}
We recover $\hat{\rho}_{jk}^{(e)}$ from this product by taking the square root and assigning the sign indicated by the fitted slope ratios.

\paragraph{Uncertainty.}
The surrogate model fit returns a covariance matrix for the fitted coefficients.
We propagate this uncertainty through the slope ratios and the symmetric estimator using the delta method, yielding a standard error $\hat{\sigma}_{\rho,jk}^{(e)}$ for each pair and environment.
These standard errors enter the ICP stage.

\subsection{Invariant causal prediction}
\label{sec:icp}

\paragraph{Environment-specific correlations.}
Each environment $e$ specifies a vector of implied patient population means $\mu^{(e)}$ and yields a set of elicited pairwise correlation estimates.
Collecting these estimates across environments lets us ask which conditional relationships remain stable across subpopulation shifts, in the spirit of ICP's observational heterogeneity framing \citep{peters2016causal}.

\paragraph{Invariance testing.}
For a target $X_t$ and a set of candidate predictors $\mathcal{C} \subseteq \mathcal{X} \setminus \{X_t\}$, ICP \citep{peters2016causal} tests subsets $S \subseteq \mathcal{C}$ for stability of the regression of $X_t$ on $S$ across environments.
If $S$ contains the direct causes of $X_t$ and the conditional mechanism is stable, these regression slopes should remain invariant.
Spurious predictors, by contrast, can change their apparent explanatory power as the marginal means shift.
For a given $(t, S)$ pair, let $\hat{R}_{SS}^{(e)}$ denote the estimated $|S|\times|S|$ predictor--predictor correlation matrix and $\hat{r}_{S,t}^{(e)}$ the vector of estimated predictor--target correlations in environment $e$. We then compute the environment-specific standardised OLS slopes as $\hat{\beta}_{t,S}^{(e)} = \bigl(\hat{R}_{SS}^{(e)}\bigr)^{-1}\hat{r}_{S,t}^{(e)}$. This is the standard OLS formula in the standardised (unit-variance, zero-mean) setting, applied directly to the elicited correlation estimates without any raw data.
Uncertainty in $\hat{\beta}_{t,S}^{(e)}$ is propagated from $\hat{\sigma}_{\rho,jk}^{(e)}$ via the delta method, yielding precision matrices $\hat{\Lambda}_{t,S}^{(e)} = [\mathrm{Var}(\hat{\beta}_{t,S}^{(e)})]^{-1}$.
We test joint invariance across all $m$ environments via the Wald statistic:
\begin{equation} \label{eq:wald_statistic}
    W_{t,S} = \sum_{e=1}^{m} \left(\hat{\beta}_{t,S}^{(e)} - \hat{\beta}_R\right)^\top \hat{\Lambda}_{t,S}^{(e)} \left(\hat{\beta}_{t,S}^{(e)} - \hat{\beta}_R\right),
\end{equation}
where $\hat{\beta}_R = \left(\sum_e \hat{\Lambda}_{t,S}^{(e)}\right)^{-1} \sum_e \hat{\Lambda}_{t,S}^{(e)} \hat{\beta}_{t,S}^{(e)}$ is the precision-weighted common slope under the null.
Under $H_0$, $W_{t,S} \sim \chi^2_{(m-1)|S|}$; we accept $S$ as invariant for target $t$ if $p > \alpha$.
After testing all subsets $S \subseteq \mathcal{C}$, we retain the variables contained in the intersection of all accepted sets.
A variable is retained only if it cannot be excluded from \emph{any} invariant set, making the estimate conservative by design.
Repeating this for each target yields a conservative set of candidate parent links.

\section{Experiments}
\label{sec:experiments}

\paragraph{Applications.}
We apply our method to two clinical domains: COPD (pneumology) and MS (neurology).
The full variable list is given in Appendix~\ref{sec:appendix-vars}, grouped by measurement modality.
All variables are expressed as z-scores (zero mean, unit variance), which puts them on a common scale despite different units and ranges. For lung function markers in COPD this matches common reporting practice and corrects for age, sex, and height.

\paragraph{Setup.}
We use OpenAI's open-weight model \texttt{gpt-oss-120b} and, for the model-size comparison, \texttt{gpt-oss-20b} \citep{openai2025gptoss}. Both models are distributed under the Apache 2.0 license, subject to the \texttt{gpt-oss} usage policy.
Prompts establish a domain-specific expert persona (pneumologist for COPD, neurologist for MS), instruct the model to consider missing values, note that measurements tend to be correlated, and require a binary choice between Patients 1 and 2.
We specify environments in the prompt as a short textual description followed by the mean values of selected variables. Unless stated otherwise, no environment is provided.

All triplet experiments use a common dense Cartesian query grid.
The core range around the center is extended symmetrically by a small margin of $0.4$ and sampled in increments of $0.2$, yielding an effective span of $\pm 1.4$ on each axis.
Patients 1 and 2 are placed symmetrically around the reference, with $X_j^{(1)} = X_{j,\mathrm{ref}} - 0.5$ and $X_j^{(2)} = X_{j,\mathrm{ref}} + 0.5$. The $X_j^{(3)}$ axis is centered at $X_{j,\mathrm{ref}}$ (default $0$ unless explicitly varied), while $X_k^{(3)}$ is centered at $0$.
Per grid point, we collect $3$ binary triplet decisions and extend to at most $6$ only if the initial responses are not unanimous. We then aggregate the patient counts and fit a binomial GLM in \texttt{statsmodels} \citep{seabold2010statsmodels} by iteratively reweighted least squares.

\paragraph{ICP.}
For ICP, we use a bipartite design: prompted environments shift a small set of clinically motivated input variables, and the selected variables serve as candidate targets.
This avoids testing links within the same modality and focuses the analysis on plausible predictor--outcome directions.
Each environment is stated explicitly in the prompt through a short clinical description together with the mean z-score values of the shifted variables.
Three environments are defined per application; their descriptions and mean z-score configurations are shown in Table~\ref{tab:icp-environments}.

\begin{table*}[t]
  \small
  \centering
  \caption{ICP environments for both applications. Each environment is described by a clinical scenario and the z-score means of the input variables (standardised z-scores).}
  \label{tab:icp-environments}
  \begin{tblr}{
    colspec = {Q[c,m,wd=0.9cm] X[l,m,font=\footnotesize] Q[c,m,wd=0.75cm] Q[c,m,wd=0.75cm] Q[c,m,wd=0.75cm]},
    hline{1,Z} = {1pt},
    hline{2} = {0.5pt},
    hline{6} = {0.5pt},
    hline{5} = {1pt},
    cell{1,5}{2} = {font=\small},
  }
    & \textit{Description} & sNfL & sGFAP & T1BHV \\
    MS1 & The patients are from a relapsing-remitting MS cohort with recent clinical relapses and evidence of active focal white matter inflammation. & +1.5 & 0.0 & -0.5 \\
    MS2 & The patients are from a progressive MS cohort with diffuse neuroinflammation and predominant grey matter involvement, without recent clinical relapses. & +0.5 & +1.5 & -0.3 \\
    MS3 & The patients are from a long-standing MS cohort with high accumulated structural lesion burden and currently low inflammatory activity. & -0.5 & -0.5 & +1.5 \\
    & \textit{Description} & pack-years & PM2.5 & BMI \\
    COPD1 & The patients are from a COPD outpatient clinic with a long history of cigarette smoking and typical urban air quality exposure. & +1.5 & 0.0 & 0.0 \\
    COPD2 & The patients are urban residents with no significant smoking history, chronically exposed to elevated ambient fine particulate matter (PM2.5). & -0.5 & +1.5 & -0.2 \\
    COPD3 & The patients are from a rural area with clean air and no significant smoking history, recruited from an obesity clinic. & -0.5 & -0.8 & +1.5 \\
  \end{tblr}
\end{table*}

\section{Results}
\label{sec:results}

\subsection{Recovered correlations}

\begin{figure}[!tbp]
  \centering
  \begin{subfigure}[t]{0.48\textwidth}
    \centering
    \includegraphics[width=\linewidth]{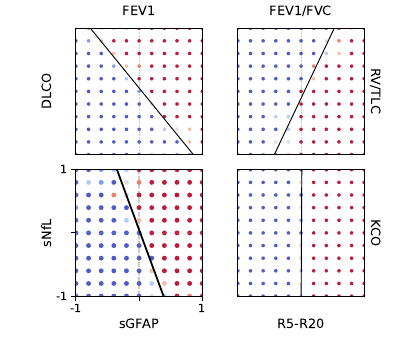}
    \caption{Triplet decision surfaces.}
    \label{fig:decision-surfaces}
  \end{subfigure}\hfill
  \begin{subfigure}[t]{0.48\textwidth}
    \centering
    \includegraphics[width=\linewidth]{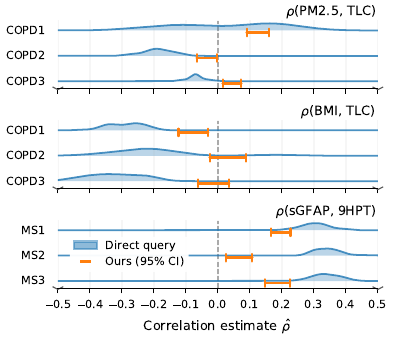}
    \caption{Direct queries and triplet estimates.}
    \label{fig:direct-rho}
  \end{subfigure}
  \caption{%
    Main extracted associations.
    (\subref{fig:decision-surfaces}) Answer-grid response fractions with surrogate decision boundaries. Axes are labelled by the clinical variable names and correspond to the queried Patient-3 values of the two variables in each panel. Blue indicates Patient~1, red indicates Patient~2 was chosen.
    (\subref{fig:direct-rho}) Direct-query distributions and triplet 95\,\% CIs.
  }
  \label{fig:results-main}
\end{figure}

\paragraph{The surrogate decision model fits well.}
Figure~\ref{fig:results-main}\subref{fig:decision-surfaces} shows the binary LLM response fractions over the queried Patient-3 value grid for four representative variable pairs.
Full directed answer-grid matrices for all COPD and MS variable pairs are provided in Appendix~\ref{sec:appendix-full-answer-grids}.
The empirical response surfaces are smooth and monotone, and the $p = 0.5$ decision boundaries estimated by logistic regression fit the decisions in all cases.
The boundary slopes are consistent with the expected sign of the correlation: for positively correlated pairs (e.g.\ \vname{FEV1} and \vname{DLCO}), the decision boundary has a negative slope---a high value of $X_k^{(3)}$ shifts the decision boundary leftward, as expected when the two variables co-vary positively.
The decision surfaces confirm that the decision pattern of the LLM is well described by the surrogate model of Section~\ref{sec:decision_surrogate}, providing empirical support for the elicitation procedure.

\paragraph{Clinical plausibility.}
Table~\ref{tab:corr-matrix-combined} reports the estimated pairwise correlations $\hat{\rho}$.
The implied correlations are clinically plausible and show clearer structure in COPD than in MS. In COPD, the strongest associations lie within measurement modalities: the spirometry variables (\vname{FEV1}, \vname{FEV1/FVC}, \vname{FEF25-75}) form a tight block ($\hat{\rho} \approx 0.68$--$0.73$), as do \vname{DLCO} and \vname{KCO} ($\hat{\rho} = 0.68$), whereas oscillometry variables (\vname{R5-R20}, \vname{X5}) are less aligned with the others, consistent with oscillometry capturing airway mechanics complementary to conventional spirometry and diffusion testing \citep{peng2025ios}. The signs of key pairs also match established physiology: \vname{FEV1/FVC} is negatively associated with \vname{RV/TLC} ($\hat{\rho} = -0.48$) and \vname{TLC} ($\hat{\rho} = -0.55$), in line with airflow obstruction, air trapping, and hyperinflation in COPD \citep{alter2020airtrapping}, while \vname{DLCO} and \vname{TLC} are positively correlated ($\hat{\rho} = 0.75$), consistent with the known relation between diffusion capacity and structural lung damage, though this relation can vary by phenotype \citep{devalla2024dlco}. In MS, correlations are weaker overall. The largest values occur in a few established pairs, such as \vname{T25FW}--\vname{T2-LV} ($\hat{\rho} = 0.40$), \vname{T25FW}--\vname{9HPT} ($\hat{\rho} = 0.33$), and \vname{sNfL}--\vname{sGFAP} ($\hat{\rho} = 0.38$), whereas most fluid-biomarker--to--clinical-scale pairs remain at or below 0.25. One plausible explanation is that cross-modality associations in MS are represented less strongly in the model, perhaps because the relevant literature is sparser or more heterogeneous than in COPD. \vname{SDMT} is particularly weakly related to the other variables, including lesion volumes, which is consistent with its use as a measure of cognitive processing speed and with the only moderate association reported between cognitive performance and white-matter lesion burden in MS \citep{morrow2025sdmt,mollison2017clinicoradiological}.

\begin{table}[t]
  \centering
  \caption{Estimated pairwise correlations for COPD and MS. Lower-left triangle: COPD. Upper-right triangle: MS. Cell color encodes the estimated correlation (orange: negative, blue: positive).}
  \label{tab:corr-matrix-combined}
  \resizebox{\linewidth}{!}{%
  \begin{tabular}{rrrrrrrrrrl}
    \textit{(a)} &  & \textcolor{gray}{{\scriptsize\shortstack[t]{sNfL}}} & {\scriptsize\shortstack[t]{sGFAP}} & {\scriptsize\shortstack[t]{EDSS}} & {\scriptsize\shortstack[t]{SDMT}} & {\scriptsize\shortstack[t]{T25FW}} & {\scriptsize\shortstack[t]{9HPT}} & {\scriptsize\shortstack[t]{T2\text{-}\\LV}} & {\scriptsize\shortstack[t]{T1\text{-}\\BHV}} & \textit{(b)} \\
    \cmidrule{3-11}
    \textcolor{gray}{{\scriptsize FEV1}} & \cellcolor{gray!20} & \cellcolor{gray!20} & \cellcolor[rgb]{0.685,0.788,0.942}{\footnotesize 0.38} & \cellcolor[rgb]{0.813,0.874,0.966}{\footnotesize 0.23} & \cellcolor[rgb]{0.997,0.978,0.960}{\footnotesize -0.04} & \cellcolor[rgb]{0.741,0.826,0.953}{\footnotesize 0.32} & \cellcolor[rgb]{0.798,0.865,0.963}{\footnotesize 0.25} & \cellcolor[rgb]{0.724,0.815,0.950}{\footnotesize 0.34} & \cellcolor[rgb]{0.851,0.900,0.973}{\footnotesize 0.18} & {\scriptsize sNfL} \\
    {\scriptsize FEV1/FVC} & \cellcolor[rgb]{0.435,0.621,0.897}{\footnotesize 0.69} & \cellcolor{gray!20} & \cellcolor{gray!20} & \cellcolor[rgb]{0.871,0.914,0.976}{\footnotesize 0.16} & \cellcolor[rgb]{0.997,0.976,0.957}{\footnotesize -0.05} & \cellcolor[rgb]{0.829,0.885,0.969}{\footnotesize 0.21} & \cellcolor[rgb]{0.866,0.910,0.975}{\footnotesize 0.16} & \cellcolor[rgb]{0.828,0.885,0.969}{\footnotesize 0.21} & \cellcolor[rgb]{0.870,0.913,0.976}{\footnotesize 0.16} & {\scriptsize sGFAP} \\
    {\scriptsize FEF25\text{-}75} & \cellcolor[rgb]{0.438,0.623,0.897}{\footnotesize 0.68} & \cellcolor[rgb]{0.401,0.598,0.890}{\footnotesize 0.73} & \cellcolor{gray!20} & \cellcolor{gray!20} & \cellcolor[rgb]{0.995,0.959,0.926}{\footnotesize -0.08} & \cellcolor[rgb]{0.820,0.880,0.967}{\footnotesize 0.22} & \cellcolor[rgb]{0.871,0.914,0.976}{\footnotesize 0.16} & \cellcolor[rgb]{0.864,0.908,0.975}{\footnotesize 0.17} & \cellcolor[rgb]{0.867,0.911,0.976}{\footnotesize 0.16} & {\scriptsize EDSS} \\
    {\scriptsize R5\text{-}R20} & \cellcolor[rgb]{0.990,0.919,0.854}{\footnotesize -0.16} & \cellcolor[rgb]{0.988,0.902,0.824}{\footnotesize -0.20} & \cellcolor[rgb]{0.993,0.945,0.901}{\footnotesize -0.11} & \cellcolor{gray!20} & \cellcolor{gray!20} & \cellcolor[rgb]{0.992,0.934,0.882}{\footnotesize -0.13} & \cellcolor[rgb]{0.999,0.988,0.979}{\footnotesize -0.02} & \cellcolor[rgb]{0.990,0.919,0.855}{\footnotesize -0.16} & \cellcolor[rgb]{0.991,0.926,0.866}{\footnotesize -0.15} & {\scriptsize SDMT} \\
    {\scriptsize X5} & \cellcolor[rgb]{0.838,0.891,0.970}{\footnotesize 0.20} & \cellcolor[rgb]{0.768,0.844,0.958}{\footnotesize 0.28} & \cellcolor[rgb]{0.778,0.851,0.959}{\footnotesize 0.27} & \cellcolor[rgb]{0.980,0.987,0.996}{\footnotesize 0.02} & \cellcolor{gray!20} & \cellcolor{gray!20} & \cellcolor[rgb]{0.727,0.817,0.950}{\footnotesize 0.33} & \cellcolor[rgb]{0.674,0.781,0.940}{\footnotesize 0.40} & \cellcolor[rgb]{0.739,0.825,0.952}{\footnotesize 0.32} & {\scriptsize T25FW} \\
    {\scriptsize DLCO} & \cellcolor[rgb]{0.411,0.605,0.892}{\footnotesize 0.72} & \cellcolor[rgb]{0.450,0.631,0.899}{\footnotesize 0.67} & \cellcolor[rgb]{0.669,0.778,0.940}{\footnotesize 0.40} & \cellcolor[rgb]{0.998,0.979,0.963}{\footnotesize -0.04} & \cellcolor[rgb]{0.860,0.906,0.974}{\footnotesize 0.17} & \cellcolor{gray!20} & \cellcolor{gray!20} & \cellcolor[rgb]{0.844,0.895,0.971}{\footnotesize 0.19} & \cellcolor[rgb]{0.901,0.933,0.982}{\footnotesize 0.12} & {\scriptsize 9HPT} \\
    {\scriptsize KCO} & \cellcolor[rgb]{0.667,0.777,0.939}{\footnotesize 0.41} & \cellcolor[rgb]{0.582,0.719,0.923}{\footnotesize 0.51} & \cellcolor[rgb]{0.750,0.832,0.954}{\footnotesize 0.31} & \cellcolor[rgb]{0.995,0.997,0.999}{\footnotesize 0.01} & \cellcolor[rgb]{0.873,0.915,0.977}{\footnotesize 0.15} & \cellcolor[rgb]{0.442,0.626,0.898}{\footnotesize 0.68} & \cellcolor{gray!20} & \cellcolor{gray!20} & \cellcolor[rgb]{0.729,0.818,0.950}{\footnotesize 0.33} & {\scriptsize T2\text{-}LV} \\
    {\scriptsize TLC} & \cellcolor[rgb]{0.546,0.695,0.917}{\footnotesize 0.55} & \cellcolor[rgb]{0.967,0.723,0.502}{\footnotesize -0.55} & \cellcolor[rgb]{0.655,0.768,0.937}{\footnotesize 0.42} & \cellcolor[rgb]{0.780,0.852,0.960}{\footnotesize 0.27} & \cellcolor[rgb]{0.867,0.911,0.976}{\footnotesize 0.16} & \cellcolor[rgb]{0.387,0.589,0.888}{\footnotesize 0.75} & \cellcolor[rgb]{0.995,0.954,0.918}{\footnotesize -0.09} & \cellcolor{gray!20} & \cellcolor{gray!20} & \textcolor{gray}{{\scriptsize T1\text{-}BHV}} \\
    {\scriptsize RV/TLC} & \cellcolor[rgb]{0.987,0.895,0.810}{\footnotesize -0.21} & \cellcolor[rgb]{0.971,0.760,0.569}{\footnotesize -0.48} & \cellcolor[rgb]{0.986,0.882,0.787}{\footnotesize -0.24} & \cellcolor[rgb]{0.761,0.840,0.956}{\footnotesize 0.29} & \cellcolor[rgb]{0.999,0.991,0.984}{\footnotesize -0.02} & \cellcolor[rgb]{0.990,0.919,0.855}{\footnotesize -0.16} & \cellcolor[rgb]{0.991,0.923,0.862}{\footnotesize -0.15} & \cellcolor[rgb]{0.646,0.762,0.935}{\footnotesize 0.43} & \cellcolor{gray!20} &  \\
    \cmidrule{1-10}
     & {\scriptsize\shortstack[t]{FEV1}} & {\scriptsize\shortstack[t]{FEV1/\\FVC}} & {\scriptsize\shortstack[t]{FEF25\text{-}\\75}} & {\scriptsize\shortstack[t]{R5\text{-}\\R20}} & {\scriptsize\shortstack[t]{X5}} & {\scriptsize\shortstack[t]{DLCO}} & {\scriptsize\shortstack[t]{KCO}} & {\scriptsize\shortstack[t]{TLC}} & \textcolor{gray}{{\scriptsize\shortstack[t]{RV/\\TLC}}} &  \\
  \end{tabular}%
  }
\end{table}

\FloatBarrier

\subsection{Invariant causal prediction}

Table~\ref{tab:icp-pvalues-pneumology} reports the ICP $p$-values for all non-empty subsets of \{\vname{pack-years}, \vname{PM2.5}, \vname{BMI}\} as candidate parents for each lung function target (COPD) and \{\vname{sNfL}, \vname{sGFAP}, \vname{T1-black-hole-volume}\} as candidate parents of the four clinical performance scales (MS).
Because ICP is conservative by design---a variable is accepted as a causal parent only if it cannot be excluded from any invariant set---the method frequently returns an empty parent set (denoted \textemdash), which should be interpreted as ``no robust causal link detected'' rather than evidence of no causal relationship.

\paragraph{COPD.}
Three links survive the intersection: \vname{pack-years} as a parent of \vname{DLCO}; \vname{PM2.5} as a parent of \vname{TLC}; and \vname{BMI} as a parent of \vname{R5-R20}.
The \vname{pack-years} $\to$ \vname{DLCO} link is clinically plausible given the established association between smoking exposure, emphysema, and reduced diffusion capacity \citep{devalla2024dlco}.
The \vname{PM2.5} $\to$ \vname{TLC} link is more tentative, but it is directionally consistent with broader evidence linking particulate air pollution to reduced lung function and increased COPD risk \citep{wang2025airpollution}.
The \vname{BMI} $\to$ \vname{R5-R20} link is also plausible, as oscillometry studies report higher respiratory impedance in obesity \citep{holtz2023obesity}. Notably, \vname{FEV1/FVC} has no accepted parent under ICP despite its clear clinical relation to smoking exposure, illustrating the conservative nature of the procedure.

\paragraph{MS.}
\vname{T1-black-hole-volume} (T1-BHV) as a parent of \vname{EDSS}, and \vname{sGFAP} as a parent of \vname{9HPT} are identified as potential associations.

The T1-BHV $\to$ \vname{EDSS} link is consistent with meta-analytic evidence that MRI lesion burden, including T1 lesion volume, is associated with disability in MS \citep{mirmosayyeb2024neuroimaging}.
The \vname{sGFAP} $\to$ \vname{9HPT} link remains exploratory, although higher sGFAP has been associated with relapse-independent disability accrual using a composite outcome that includes \vname{9HPT} \citep{rosenstein2024sgfap}.

\begin{table*}[t]
  \centering
  \makebox[\textwidth][c]{\begingroup
\setlength{\tabcolsep}{0pt}
\begin{tabular}{@{}>{\centering\arraybackslash}m{1.9em}@{\hspace{3.8pt}}>{\centering\arraybackslash}m{1.9em}@{\hspace{3.8pt}}>{\centering\arraybackslash}m{1.9em}@{\hspace{3pt}}>{\raggedleft\arraybackslash}m{2.5em}@{\hspace{3.8pt}}>{\raggedleft\arraybackslash}m{2.5em}@{\hspace{3.8pt}}>{\raggedleft\arraybackslash}m{2.5em}@{\hspace{3.8pt}}>{\raggedleft\arraybackslash}m{2.5em}@{\hspace{3.8pt}}>{\raggedleft\arraybackslash}m{2.5em}@{\hspace{1.8em}}>{\centering\arraybackslash}m{1.9em}@{\hspace{3.8pt}}>{\centering\arraybackslash}m{1.9em}@{\hspace{3.8pt}}>{\centering\arraybackslash}m{1.9em}@{\hspace{3pt}}>{\raggedleft\arraybackslash}m{2.5em}@{\hspace{3.8pt}}>{\raggedleft\arraybackslash}m{2.5em}@{\hspace{3.8pt}}>{\raggedleft\arraybackslash}m{2.5em}@{\hspace{3.8pt}}>{\raggedleft\arraybackslash}m{2.5em}@{}}
    \toprule
    \multicolumn{8}{c}{COPD} & \multicolumn{7}{c}{MS} \\
    \cmidrule(lr){1-8}\cmidrule(lr){9-15}
    {\scriptsize \shortstack{pack-\\years}} & {\scriptsize PM2.5} & {\scriptsize BMI} & {\scriptsize \shortstack{FEV1/\\FVC}} & {\scriptsize RV/TLC} & {\scriptsize TLC} & {\scriptsize DLCO} & {\scriptsize R5-R20} & {\scriptsize sNfL} & {\scriptsize sGFAP} & {\scriptsize T1BHV} & {\scriptsize EDSS} & {\scriptsize SDMT} & {\scriptsize T25FW} & {\scriptsize 9HPT} \\
    \midrule
    $\circ$ & $\circ$ & $\bullet$ & {\small \textbf{.000}} & {\small .989} & {\small \textbf{.004}} & {\small \textbf{.000}} & {\small .056} & $\circ$ & $\circ$ & $\bullet$ & {\small .579} & {\small .631} & {\small \textbf{.000}} & {\small \textbf{.000}} \\
    $\circ$ & $\bullet$ & $\circ$ & {\small \textbf{.002}} & {\small \textbf{.000}} & {\small .052} & {\small \textbf{.000}} & {\small \textbf{.002}} & $\circ$ & $\bullet$ & $\circ$ & {\small \textbf{.048}} & {\small .987} & {\small \textbf{.000}} & {\small .539} \\
    $\bullet$ & $\circ$ & $\circ$ & {\small \textbf{.003}} & {\small .982} & {\small \textbf{.038}} & {\small .720} & {\small \textbf{.048}} & $\bullet$ & $\circ$ & $\circ$ & {\small \textbf{.006}} & {\small .830} & {\small \textbf{.026}} & {\small \textbf{.000}} \\
    $\circ$ & $\bullet$ & $\bullet$ & {\small \textbf{.000}} & {\small \textbf{.010}} & {\small \textbf{.001}} & {\small \textbf{.000}} & {\small \textbf{.000}} & $\circ$ & $\bullet$ & $\bullet$ & {\small .183} & {\small .927} & {\small \textbf{.000}} & {\small \textbf{.001}} \\
    $\bullet$ & $\circ$ & $\bullet$ & {\small \textbf{.000}} & {\small .983} & {\small \textbf{.004}} & {\small \textbf{.000}} & {\small \textbf{.003}} & $\bullet$ & $\circ$ & $\bullet$ & {\small .052} & {\small .848} & {\small \textbf{.000}} & {\small \textbf{.000}} \\
    $\bullet$ & $\bullet$ & $\circ$ & {\small \textbf{.000}} & {\small \textbf{.002}} & {\small .061} & {\small \textbf{.000}} & {\small \textbf{.015}} & $\bullet$ & $\bullet$ & $\circ$ & {\small \textbf{.001}} & {\small .965} & {\small \textbf{.000}} & {\small \textbf{.002}} \\
    $\bullet$ & $\bullet$ & $\bullet$ & {\small \textbf{.000}} & {\small .077} & {\small \textbf{.006}} & {\small \textbf{.000}} & {\small \textbf{.001}} & $\bullet$ & $\bullet$ & $\bullet$ & {\small \textbf{.014}} & {\small .958} & {\small \textbf{.000}} & {\small \textbf{.000}} \\
    \midrule
    \multicolumn{3}{c}{\textit{parents}} & {\scriptsize \shortstack[c]{\textemdash{}}} & {\scriptsize \shortstack[c]{\textemdash{}}} & {\scriptsize \shortstack[c]{PM2.5}} & {\scriptsize \shortstack[c]{\shortstack{pack-\\years}}} & {\scriptsize \shortstack[c]{BMI}} & \multicolumn{3}{c}{\textit{parents}} & {\scriptsize \shortstack[c]{T1BHV}} & {\scriptsize \shortstack[c]{\textemdash{}}} & {\scriptsize \shortstack[c]{\textemdash{}}} & {\scriptsize \shortstack[c]{sGFAP}} \\
    \bottomrule
\end{tabular}
\endgroup}
  \caption{ICP p-values for COPD and MS targets.
    Filled circles indicate included candidate parents; open circles indicate excluded ones.
    \textbf{Bold} entries are rejected at $p < 0.05$.}
  \label{tab:icp-pvalues}
  \label{tab:icp-pvalues-pneumology}
  \label{tab:icp-pvalues-ms}
\end{table*}

\subsection{Validation and sensitivity analyses}

\begin{figure}[!tbp]
  \centering
  \begin{subfigure}[t]{0.48\textwidth}
    \centering
    \includegraphics[width=\linewidth]{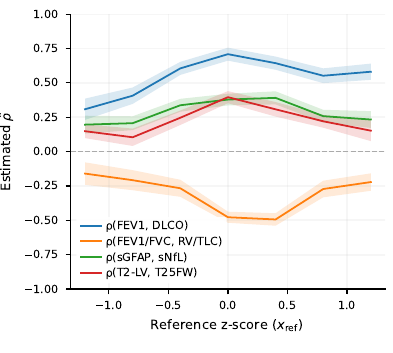}
    \caption{Sensitivity to reference choice.}
    \label{fig:xref-sweep}
  \end{subfigure}\hfill
  \begin{subfigure}[t]{0.48\textwidth}
    \centering
    \includegraphics[width=\linewidth]{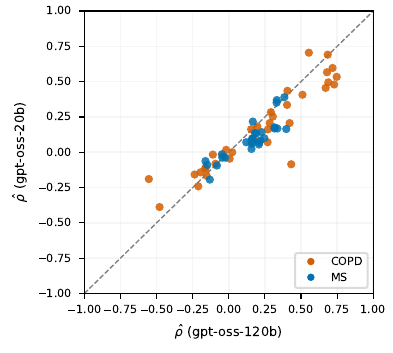}
    \caption{Model-size comparison.}
    \label{fig:model-scatter}
  \end{subfigure}
  \caption{%
    Robustness checks.
    (\subref{fig:xref-sweep}) Sensitivity to the reference value $X_{j,\mathrm{ref}}$ for the queried first variable.
    (\subref{fig:model-scatter}) Model-size comparison of pairwise $\hat{\rho}$ estimates.
  }
  \label{fig:results-robustness}
\end{figure}

\paragraph{Comparison with direct queries.}
Figure~\ref{fig:results-main}\subref{fig:direct-rho} compares the distributions obtained by asking the LLM directly for Pearson correlations (blue kernel density estimates, KDE, based on 50 repetitions) with the estimates from our triplet method (orange 95\,\% CIs).
Direct-query responses are much more dispersed, with standard deviations typically in the range $0.1$--$0.3$, whereas the triplet-based confidence intervals are narrower; for $\hat{\rho}_{\mathrm{PM2.5},\mathrm{TLC}}$, the direct-query distribution is even bimodal, with one positive and one negative peak. They are also less sensitive to prompted environments: the distributions overlap substantially across environments for most variable pairs, while the triplet estimates vary more strongly, as required for ICP. 
Taken together, these patterns suggest that triplet-based correlation estimation is less noisy than direct queries, especially when it comes to discerning differences across environments.

\paragraph{Reference choice.}
Figure~\ref{fig:results-robustness}\subref{fig:xref-sweep} shows how $\hat{\rho}$ varies as a function of the reference value $X_{j,\mathrm{ref}}$ for selected variable pairs.
Estimates are stable in the central range ($|X_{j,\mathrm{ref}}| \lesssim 1$) and tend to shrink toward zero in the tails, as expected when the anchor and reference patients become more similar and implied differences are harder to distinguish.
One exception is the \vname{FEV1}--\vname{DLCO} pair in COPD, which remains elevated even at positive $X_{j,\mathrm{ref}} \approx 1$.

\paragraph{Model size.}
Figure~\ref{fig:results-robustness}\subref{fig:model-scatter} compares pairwise $\hat{\rho}$ estimates from the large model (\texttt{gpt-oss-120b}, horizontal axis) and the smaller model (\texttt{gpt-oss-20b}, vertical axis) across both applications.
Most points lie close to the diagonal, indicating that both models recover a similar relative correlation structure across variable pairs.
The \texttt{gpt-oss-20b} estimates are, however, slightly smaller in absolute magnitude, suggesting that the smaller model has internalized the correlation structure less strongly even though the overall association pattern remains similar.

\section{Related Work}
\label{sec:related_work}

\paragraph{LLMs as sources of causal knowledge.}
Some authors queried models directly about causal relations and compared the answers with known graphs or benchmark tasks \citep{long2023can,jin2023can,jin2023cladder,zhou2024causalbench,zecevic2023causal}.
Others used LLM judgments as priors, constraints, or refinements for statistical causal discovery, rather than as complete discovery procedures \citep{long2023causal,ban2023causal,ban2023integrating,darvariu2024large,takayama2024integrating}.
\citet{wan2025large} survey this literature and distinguish direct inference, prior-knowledge integration, and post-hoc refinement.
Most of this work elicits knowledge through explicit statements about edges, directions, or causal relevance.
We instead recover dependence structure via comparison questions that do not rely on the quality of quantitative statements from LLMs and apply invariant causal prediction only as a second step \citep{peters2016causal}.

\paragraph{Knowledge probing and behavioral elicitation.}
The LAMA benchmark tested whether a model can recover knowledge-base triples from cloze queries \citep{petroni2019language}, and later surveys organized a broad set of related probing methods and datasets \citep{youssef2023give}.
Follow-up work showed that the answer depends strongly on the prompt used to elicit the fact \citep{shin2020autoprompt}.
Biomedical variants extend this setting to clinical and scientific relations, where multi-token entities, long-tailed vocabularies, and many-to-many relations make probing harder \citep{sung2021can,meng2022rewirethenprobe,yao2023context}.
These studies mostly targeted discrete facts such as entity--relation triples.
Closer to our setting, \citet{requeima2024llm} elicited numerical predictive distributions from LLMs conditioned on natural-language descriptions and numerical context.
Instead of asking the model directly, we infer continuous associations using triplet-style comparisons, which have been used to recover similarity structure from qualitative judgments \citep{vankadara2023insights} and to test whether clinical embeddings agree with expert similarity judgments \citep{kabus2026assessing}.

\paragraph{Probing and mechanistic interpretability.}
Probing classifiers can reveal whether representations contain linguistic, factual, spatial, temporal, or truth-related information \citep{belinkov2021probing,gurnee2024language,marks2024geometry,mallen2024eliciting}.
However, encoded information need not be used by the model in the behavior of interest.
This distinction has motivated stronger causal tests, including amnesic probing, mediation analysis, activation patching, and model editing \citep{elazar2021amnesic,vig2020investigating,meng2023locating,conmy2023automated,lindsey2025biology}.
These methods can provide mechanistic evidence, but they require access to internal activations and often assume that the relevant computation can be localized.
We instead study the model behaviorally, using controlled prompts and observed choices rather than internal activations.

\section{Conclusion}
\label{sec:conclusion}

We introduced a behavioral framework for extracting association structure from LLMs using structured triplet comparisons. The method turns binary similarity judgments into pairwise correlation estimates without requiring access to model internals, uses prompted subpopulation shifts to compare these estimates across environments and applies ICP as a conservative filter for candidate parent links. 
In COPD and MS, the recovered correlations were stable and clinically interpretable, with a stronger and more coherent signal in COPD than in MS. Compared with direct correlation queries, the triplet-based estimates were less dispersed and more sensitive to prompted environment shifts, which is important for downstream invariance testing. 
Additional robustness checks showed stable estimates over central reference choices and broadly consistent association structure across model sizes. These results suggest that structured comparison queries can expose meaningful association structure represented in LLMs.

Several limitations qualify this interpretation.
Most notably, the method recovers model-implied rather than empirical clinical associations.
Moreover, although the surrogate model fits the observed triplet decisions well, the translation from fitted slopes to correlation estimates still depends on the relational assumptions underlying the estimator and may not capture non-linear internal structure.
Finally, the empirical study is limited to two clinical domains, a modest variable set, and a small set of models.
Within these limits, the proposed framework provides an exploratory tool for probing association structure in LLMs.

\begin{ack}
Funded by the Deutsche Forschungsgemeinschaft (DFG, German Research Foundation) -- Project-ID 499552394 -- SFB 1597.
\end{ack}

\bibliographystyle{plainnat}
\bibliography{manual}

\clearpage
\appendix
\onecolumn

\section{Prompt Templates}
\label{sec:appendix-prompts}

This appendix shows the abstract prompt templates used in the experiments.
We replace implementation-specific fields by placeholders such as
\texttt{[expert persona]}, \texttt{[environment description]}, and the variable
names $X_j$ and $X_k$, while keeping the substantive wording of the prompts
unchanged. Here \texttt{[environment description]} stands for the full
environment specification: the textual cohort description together with the mean
values of the shifted variables, as listed in Table~\ref{tab:icp-environments}.
All numeric values are rendered with one decimal place.

\begin{samepage}
\noindent\textbf{Triplet comparison prompt.}
\begin{quote}
\ttfamily\small
[expert persona]\par
All values are population-normalized z-scores: 0.0 = population mean, 1.0 =
one population standard deviation.\par
[environment description]\par
\vspace{0.5\baselineskip}
Patient 1: $X_j$ = [value], $X_k$ = missing\par
Patient 2: $X_j$ = [value], $X_k$ = missing\par
Patient 3: $X_j$ = [value], $X_k$ = [value]\par
\vspace{0.5\baselineskip}
Infer the missing $X_k$ values for Patients 1 and 2 from their observed $X_j$
values.\par
Be mindful that clinical variables and biomarkers tend to be highly
correlated.\par
Then choose which patient (1 or 2) is more similar to Patient 3.\par
\vspace{0.5\baselineskip}
Respond with exactly one line containing only:\par
1\par
or\par
2
\end{quote}
\end{samepage}

\begin{samepage}
\noindent\textbf{Direct correlation query prompt.}
\begin{quote}
\ttfamily\small
[expert persona]\par
\vspace{0.5\baselineskip}
What is the Pearson correlation coefficient between $X_j$ and $X_k$ in the
relevant patient population?\par
[environment description]\par
\vspace{0.5\baselineskip}
Reason through this carefully. Consider the physiological or clinical
relationship between the two variables, draw on specific studies, cohorts, or
reference ranges you are aware of, and discuss any factors that might
strengthen or attenuate the association. Weigh conflicting evidence if it
exists.\par
\vspace{0.5\baselineskip}
On the very last line of your response, write exactly (and nothing else):\par
correlation: X.XX
\end{quote}
\end{samepage}

\noindent\textbf{Persona texts.}
\begin{center}
  \captionsetup{type=table}
  \captionof{table}{Prompt persona texts.}
  \label{tab:appendix-personas}
  \centering
  \begin{tabularx}{\linewidth}{@{}lX@{}}
    \toprule
    Domain & Persona text \\
    \midrule
    COPD & You are a pneumologist specialized in obstructive and restrictive lung disease. \\
    MS & You are a neurologist specialized in multiple sclerosis and demyelinating disease. \\
    \bottomrule
  \end{tabularx}
\end{center}

\clearpage

\section{Variable Reference}
\label{sec:appendix-vars}

\begin{center}
  \small
  \captionsetup{type=table}
  \captionof{table}{Variables and descriptions for COPD and MS.}
  \label{tab:vars-appendix}
  \begin{tblr}{
    colspec = {Q[c,m,wd=0.55cm] Q[l,m,wd=2.4cm] l X},
    hline{1,Z} = {1pt},
    hline{2} = {0.5pt},
    hline{11} = {1pt},
    hline{5,7,9} = {2-4}{wd=0.3pt, fg=gray!50},
    hline{13,17} = {2-4}{wd=0.3pt, fg=gray!50},
    row{10} = {belowsep=4pt},
    row{11} = {abovesep=4pt},
    column{2} = {font=\itshape},
  }
    & \textbf{Modality} & \textbf{Variable} & \textbf{Description} \\
    \SetCell[r=9]{c,m} \rotatebox{90}{COPD}
      & \SetCell[r=3]{l,m} Spirometry
        & FEV1     & Forced expiratory volume in 1\,s. \\
    & & FEV1/FVC & Ratio of FEV1 to FVC; obstructive ratio. \\
    & & FEF25-75 & Mean forced expiratory flow, 25\%--75\% of FVC. \\
    & \SetCell[r=2]{l,m} Oscillometry
        & R5-R20   & Frequency-dependent resistance (5\,Hz $-$ 20\,Hz). \\
    & & X5       & Respiratory system reactance at 5\,Hz. \\
    & \SetCell[r=2]{l,m} Diffusion
        & DLCO     & Diffusing capacity for carbon monoxide. \\
    & & KCO      & Transfer coefficient (DLCO / alveolar volume). \\
    & \SetCell[r=2]{l,m} Plethysmography
        & TLC      & Total lung capacity. \\
    & & RV/TLC   & Residual volume fraction of total lung capacity. \\
    \SetCell[r=8]{c,m} \rotatebox{90}{MS}
      & \SetCell[r=2]{l,m} Fluid Biomarkers
        & sNfL     & Neurofilament light chain; neuroaxonal damage. \\
    & & sGFAP    & Glial fibrillary acidic protein; astrocyte damage. \\
    & \SetCell[r=4]{l,m} Clinical Scales
        & EDSS     & Expanded Disability Status Scale. \\
    & & SDMT     & Symbol Digit Modalities Test; processing speed. \\
    & & T25FW    & Timed 25-Foot Walk; ambulation. \\
    & & 9HPT     & 9-Hole Peg Test; manual dexterity. \\
    & \SetCell[r=2]{l,m} MRI Lesions
        & T2-lesion-volume     & T2-hyperintense white matter lesion volume. \\
    & & T1-black-hole-volume & T1-hypointense lesion volume; axonal loss. \\
  \end{tblr}
\end{center}

\clearpage

\section{Full Answer Grids}
\label{sec:appendix-full-answer-grids}

Figures~\ref{fig:appendix-answer-grid-copd} and~\ref{fig:appendix-answer-grid-ms}
show the full directed answer-grid matrices for all COPD and MS variable pairs.

\vspace{0.75\baselineskip}

\begin{center}
  \centering
  \IfFileExists{generated/answer_grid/answer_grid_copd.pdf}{%
    \makebox[\textwidth][c]{%
      \includegraphics[width=0.98\textwidth,height=0.71\textheight,keepaspectratio]{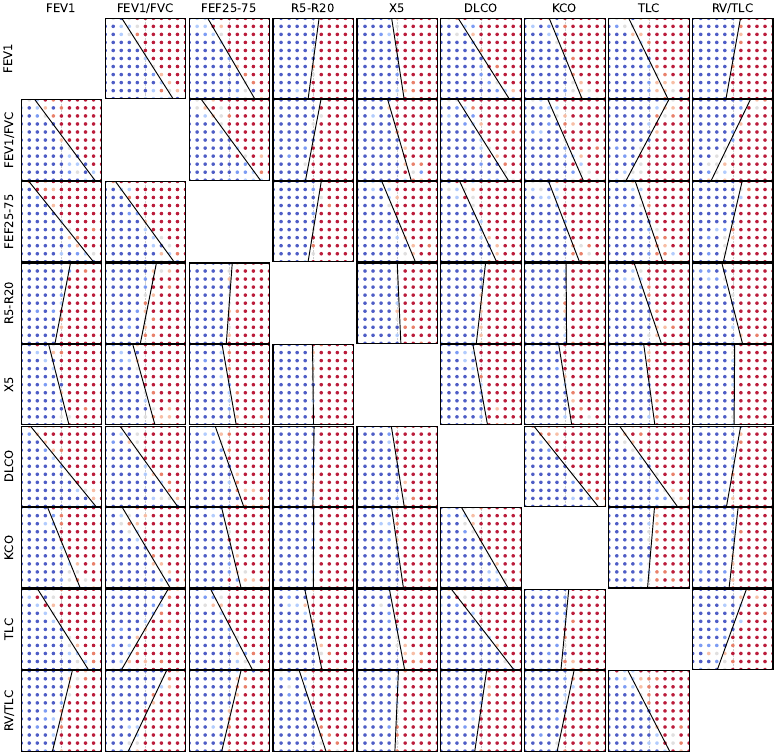}%
    }%
  }{%
    \fbox{\parbox{0.9\linewidth}{\centering
    Full COPD answer-grid figure not generated yet.\\
    Run \texttt{just paper\_answer\_grid} in the \texttt{llm-causal} repository to create
    \texttt{generated/answer\_grid/answer\_grid\_copd.pdf}.}}%
  }
  \captionsetup{type=figure}
  \captionof{figure}{Full COPD answer-grid matrix.}
  \label{fig:appendix-answer-grid-copd}
\end{center}

\clearpage
\vspace*{0.03\textheight}
\begin{center}
  \centering
  \IfFileExists{generated/answer_grid/answer_grid_ms.pdf}{%
    \makebox[\textwidth][c]{%
      \includegraphics[width=0.98\textwidth,height=0.78\textheight,keepaspectratio]{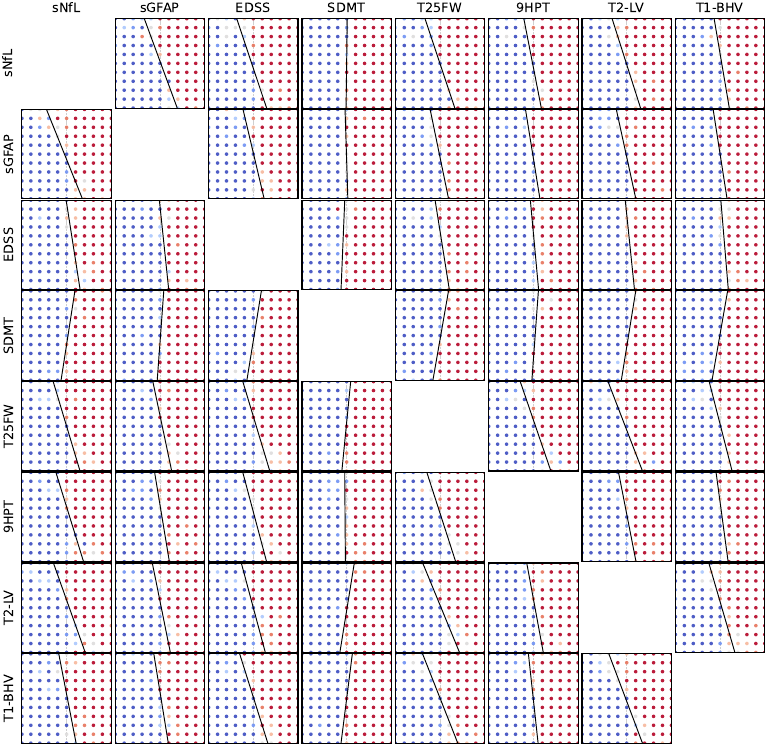}%
    }%
  }{%
    \fbox{\parbox{0.9\linewidth}{\centering
    Full MS answer-grid figure not generated yet.\\
    Run \texttt{just paper\_answer\_grid} in the \texttt{llm-causal} repository to create
    \texttt{generated/answer\_grid/answer\_grid\_ms.pdf}.}}%
  }
  \captionsetup{type=figure}
  \captionof{figure}{Full MS answer-grid matrix.}
  \label{fig:appendix-answer-grid-ms}
\end{center}

\clearpage
\section{Compute Resources}
\label{sec:appendix-compute-resources}

Table~\ref{tab:appendix-request-counts} summarizes the number of LLM
requests required by the main analysis blocks underlying the reported results.
The experiments were run on our local cluster. For the Slurm jobs underlying
the reported artifacts, we reserved 8 CPU cores, 48\,GB host memory, and 1 GPU.
The model server ran via vLLM on a single NVIDIA H100 GPU with
95{,}830\,MiB VRAM; the compute node class uses AMD EPYC 9454 processors. 
After warm-up, average throughput was approximately 9 requests
per second. During method development, some analyses were rerun multiple times,
so total development-time request counts were higher than the final counts
reported here, although caching reduced repeated inference costs.

\begin{center}
  \captionsetup{type=table}
  \captionof{table}{LLM request counts for the main analysis blocks.}
  \label{tab:appendix-request-counts}
  \centering
  \renewcommand{\arraystretch}{1.2}
  \begin{tabularx}{\textwidth}{@{}lXr@{}}
    \toprule
    Analysis block & Supported artifacts & Requests \\
    \midrule
    \shortstack[l]{Pairwise correlation (\texttt{gpt-oss-120b})} & Table~\ref{tab:corr-matrix-combined} and Figure~\ref{fig:results-main}\subref{fig:decision-surfaces} & 52,545 \\
    \addlinespace[2pt]
    \shortstack[l]{Pairwise correlation (\texttt{gpt-oss-20b})} & Figure~\ref{fig:results-robustness}\subref{fig:model-scatter} & 53,400 \\
    \addlinespace[2pt]
    ICP analysis & Table~\ref{tab:icp-pvalues} and Figure~\ref{fig:results-main}\subref{fig:direct-rho} & 78,960 \\
    \addlinespace[2pt]
    Reference-value sweeps & Figure~\ref{fig:results-robustness}\subref{fig:xref-sweep} & 23,871 \\
    \bottomrule
  \end{tabularx}
\end{center}

\end{document}